\documentclass[letterpaper, 10 pt, conference]{ieeeconf}  

\IEEEoverridecommandlockouts                              

\usepackage{amsmath} 
\usepackage{amssymb}  
\usepackage{dsfont}
\usepackage{graphicx}
\usepackage{booktabs}
\usepackage{psfrag,graphicx,epsfig}
\usepackage{epstopdf}
\usepackage{xspace}
\usepackage{subfig}
\usepackage{float}
\usepackage{placeins}
\usepackage{multirow}
\usepackage{pgf,tikz}
\usepackage{nowidow}
\usepackage{lineno}
\usepackage{xcolor}

\usepackage{siunitx}
\usepackage{color}
\usepackage{flushend}
\usepackage{lineno}
\usepackage{algorithmic}
\usepackage{float}
\usepackage{mathtools}
\allowdisplaybreaks

\title{\LARGE \bf
Learning Near-global-optimal Strategies for Hybrid Non-convex Model Predictive Control of Single Rigid Body Locomotion 
}

\author{Xuan Lin$^{\dagger}$, Feng Xu$^{\dagger}$, Alexander Schperberg$^{\dagger}$, and Dennis Hong$^{\dagger}$%
\thanks{$^{\dagger}$ All authors are with the Department of Mechanical and Aerospace Engineering, University of California, Los Angeles, CA, USA 90095 {\tt\small \{maynight, xufengmax, aschperberg28, dennishong\}@g.ucla.edu}}}%

\begin{document}
\bstctlcite{IEEEexample:BSTcontrol}

\maketitle


\begin{abstract}
Convex model predictive controls (MPCs) with a single rigid body model have demonstrated strong performance on real legged robots. However, convex MPCs are limited by their assumptions such as small rotation angle and pre-defined gait, limiting the richness of potential solutions. We remove those assumptions and solve the complete mixed-integer non-convex programming with single rigid body model. We first collect datasets of pre-solved problems offline, then learn the problem-solution map to solve this optimization fast for MPC. If warm-starts can be found, offline problems can be solved close to the global optimality. The proposed controller is tested by generating various gaits and behaviors depending on the initial conditions. Hardware test demonstrates online gait generation and adaptation running at more than 50 Hz based on sensor feedback.

\end{abstract}

\section{Introduction}\label{sec:introduction}
Leg robots have demonstrated advantageous traversability over uneven or discrete terrains compared to wheeled robots. Legged robots achieve such an advantage by making discrete contact with the ground while keeping a balance of its body. This enables it to walk over stairs \cite{siekmann2021blind}, jumping within complex environment \cite{yim2018precision}, or climb between multiple walls \cite{lin2018multi}. It is thus essential to design a controller that maintains the robot body balance in the presence of external disturbances. Two of the currently most popular approaches are model-based methods such as model predictive control \cite{di2018dynamic}, and learning-based methods such as reinforcement learning \cite{hwangbo2019learning}. 

The basic idea of model predictive control is to solve online several steps control signals into the future given the sensor feedback but only execute the first step control signal, then solve the problem again. This allows the controller to dynamically adjust its output according to the feedback. To solve the control signals, optimization-based formulations are usually used online. This controller is model-based, hence if the model is complicated, the solving speed can be too slow to catch up with the natural dynamics of robots. As as result, simplified models such as linearization are usually used \cite{di2018dynamic}. On the other hand, reinforcement learning has shown good performance for stabilizing the robot body for quadruped robots \cite{hwangbo2019learning} and bipedal robots \cite{siekmann2021blind}. Reinforcement learning does not require a model and relies on simulated environments to collect a large amount of data for training robust policies. 

One of the biggest challenges of legged locomotion for model-based controllers is to let the controller make or break contacts with the environment while sustaining a continuous dynamic system. Previous literature has adopted several approaches to address such a hybrid system control problem \cite{kong2021ilqr} \cite{cleac2021fast}. There are in general two approaches for optimization over discrete variables: mixed-integer optimization formulation and complementary formulation. However, those approaches are usually computationally expensive and not tractable online, limiting their applications to non-time-critical motion planning or in some cases, solving small-scale problems for control \cite{cleac2021fast}. However, recent progress has been made to incorporate learning methods to discover heuristics from data and accelerate the solver, pushing relatively larger scale problems towards control applications \cite{lin2021reduce}.

In this paper, we build upon the previous work \cite{lin2022benchmark} and propose a framework to use mixed-integer non-convex programming for model predictive control for quadruped locomotion problems. The dynamics are non-convex and not linearized, the gait is chosen online, and the body orientation change is not limited to small angles. Consequently, the problem formulation incorporates non-convex multi-linear terms for dynamics and binary terms for gait selection, hence a mixed-integer non-convex optimization problem (MINLP). In the previous paper \cite{lin2021reduce}, we have compared several equivalent approaches to resolve this problem including the nonlinear alternating direction method of multipliers (ADMM) approach \cite{shirai2022simultaneous}, the data-driven approach to warm-start mixed-integer programs or mathematical programs with complementary constraints online, and the data-driven approach to solve the convex programmings online \cite{cauligi2021coco}. The convex programming approach shows potential to solve the problem with several trials of convex programming given well-performed supervised learning. The solving speed of this approach is the fastest due to powerful convex optimization solvers such as OSQP \cite{stellato2020osqp}. We adopt this approach to resolve our mixed-integer nonlinear programming problems online for MPC.

In general, the multi-body dynamics of robots such as manipulators incorporate a large number of multi-linear terms. Creating convex relaxations of those models is non-trivial. One naive approach is to break the multi-linearity into several bilinear constraints and use McCormick envelopes to approximate them. However, for multi-linear constraints with high degree, this approach has low accuracy. Fortunately, for single rigid body models, the degree of multi-linearity is tractable and we can write convex relaxations using McCormick envelope constraints with reasonable accuracy, and convert the non-convex constraints into mixed-integer envelope constraints. Envelope constraints are usually very slow to solve. Offline, we sample a number of problem instances from a distribution and pre-solve those problems. A learner is then trained to provide integer variables given a problem received online. With learned warm-start integers, the problem can be converted into convex programming permitting a convex solver online for control input. 

There exist several numerical approaches such as ADMM \cite{shirai2022simultaneous}, sequential convex programmings \cite{quirynen2021sequential}, and so on. However, those methods focus on getting locally optimal solutions sacrificing optimality. Additionally, their solving speed is usually slower. The merit of data-driven methods is that since the dataset is collected offline, we can iterate the dataset over the nonlinear programming a few times (use the result as warm-starts for the next iteration). This will improve the optimality of the dataset. Finally, we can push the dataset through the equivalent MIP formulation and solve to near global optimal solutions. In this paper, we can limit the gap between most of the current solutions to the theoretically best solution to lower than 15\% through the Gurobi solver. We use the proposed control framework to generate trajectories for a quadruped robot walking and jumping under different scenarios, and implement the framework on the actual hardware to solve a walking with disturbance rejection problem with gait adaptation online. 

To summarize, our contributions are as follows:
\begin{enumerate}
    \item We scale up the methods of learning problem-solution map to more realistic model predictive control problems incorporating mixed-integer non-convex constraints and push the solving speed for online applications.
    \item We demonstrate the proposed controller on actual hardware.
\end{enumerate}


\section{Dynamic Model}\label{sec:dynamics}
In this section, we explain the dynamic model used for control in this paper. Following the common approach seen in literature, we model the robot into a single rigid body ignoring the swing leg dynamics. However, we do not make any simplification of the model such as the small angle approximation which is commonly seen in the literature \cite{di2018dynamic}. We also do not use a pre-planned gait.

\subsection{Nonlinear Dynamics}

The dynamics of a single rigid body including the Newton second law and the angular momentum are:

\begin{equation}
    m\ddot{\textbf{p}} = \sum_{i=1}^{N}\textbf{f}_{i}-m\textbf{g}
\end{equation}

\begin{equation}
\textbf{I}\dot{\boldsymbol{\omega}} + \boldsymbol{\omega} \times (\textbf{I}\boldsymbol{\omega}) = \sum_{i=1}^{n} \textbf{r}_{b,i} \times \textbf{f}_{i}   
\end{equation}

Where $\textbf{p}$ represents the geometric center of the body which is also assumed to be the center of mass. $\textbf{f}_{i}$ represents the contact force on toe $i$. $\boldsymbol{\omega}$ is the angular velocity and $\textbf{I}$ is the moment of inertia. $\textbf{r}_{b,i}$ is the vector from body center of mass position $\textbf{p}$ to the contact position $\textbf{p}_{i}$ which is the moment arm of $\textbf{f}_{i}$. 


We define the Z-Y-X Euler angles as $\boldsymbol{\Theta}=[\phi, \theta, \psi]$. The transformation matrix from body frame to world frame is:

\begin{equation}
\textbf{R}_{wb} = 
\begin{bmatrix}
c \theta c \psi & s \phi s \theta c \psi  - s \psi c \phi  & s \phi s \psi  + s\theta c \phi c \psi  \\
s \psi c \theta & c \phi c \psi  + s \phi s \theta s \psi & s \theta s \psi c \phi  - s\phi c \psi  \\
-s \theta & s\phi c \theta & c \phi c \theta
\end{bmatrix}
\end{equation}

The transformation from the rate of change of Euler angles to angular velocities $\boldsymbol{\omega}$ is:
\begin{equation}
\boldsymbol{\omega} = 
\begin{bmatrix}
c \theta c \psi  & -s \psi  & 0 \\
c \theta s \psi  & c \psi  & 0 \\
-s \theta  & 0 & 1
\end{bmatrix}
\begin{bmatrix}
\dot{\phi} \\
\dot{\theta} \\
\dot{\psi} \\
\end{bmatrix}
\label{Eqn:omega}
\end{equation}

In this paper, we assume that the robot is making point contact to the ground. The friction cone constraint on each toe can be written as:

\begin{equation}
    f_{i,z} \geq \mu \sqrt{ f_{i,x}^{2} + f_{i,y}^{2} }
\end{equation}

Where $\mu$ is the coefficient of friction. Following the common approaches of trajectory optimization, the continuous trajectory is discretized into knot points at step $1, 2, ..., N$ with fixed $\Delta T$. Between knot points, the forward Euler integration constraint is enforced:
\begin{equation}
\begin{aligned}
\textbf{p}[n+1] - \textbf{p}[n] = \textbf{v} \Delta T \\
\textbf{v}[n+1] - \textbf{v}[n] = \textbf{a} \Delta T \\
\boldsymbol{\Theta}[n+1] - \boldsymbol{\Theta}[n] = \boldsymbol{\Theta} \Delta T \\
\boldsymbol{\omega}[n+1] - \boldsymbol{\omega}[n] = \boldsymbol{\alpha} \Delta T \\
\end{aligned}
\end{equation}

Where $\textbf{v}$ is the center of mass velocity and $\textbf{a}$ is the center of mass acceleration. $\boldsymbol{\alpha}$ is the body angular acceleration.

The relation between the vector of toe position $\textbf{p}_{w,i}$ originated from world frame origin to the vector of toe position $\textbf{p}_{b,i}$ originated from the body center of mass is: 

\begin{equation}
    \textbf{p}_{w,i}[n+1] - \textbf{p}[n] = \textbf{p}_{b,i}[n+1]
\end{equation}

For easiness of solving, we also make approximation of the leg workspace into a box:

\begin{equation}
\begin{aligned}
    \textbf{p}[n] + \textbf{R}_{wb}[n] \textbf{H}_{b, i} & = \textbf{H}_{w, i}[n] \\
    \textbf{p}_{w,i}[n] - \textbf{H}_{w, i}[n] - \textbf{R}_{wb}[n] o_{i} & \in \textbf{R}_{wb}[n]B
\end{aligned}
\end{equation}
                   
Where $o_{i}$ is the offset from body center to the shoulder. $\textbf{H}_{b, i}$ is the vector from the robot center of mass to the shoulder position (base position of leg) of the robot body. $\textbf{H}_{w, i}$ is the vector from the origin of the world frame to the shoulder position. $B$ is the size of box for leg workspace approximation. Note that since we collect data offline, we can use full kinematics to completely explore the workspace. Online, we can discretize the workspace into convex regions and formulate the kinematics as mixed-integer convex constraint \cite{lin2019optimization}.

In order to characterize gait, we define contact variable $c_{i}$ for each toe. If $c_{i}=0$, the leg $i$ is lifted to the air. If $c_{i}=1$, the leg is down on the ground. Since each toe can only have one state, $c_{i}$ is a binary variable.

In addition, the no-slip condition is enforced:
\begin{equation}
    |\textbf{p}_{w,i}[n+1] - \textbf{p}_{w,i}[n]| \leq (1-c_{i}[n])M + (1-c_{i}[n+1])M 
\end{equation}

Where $M$ is the standard bigM constant. This mixed-integer linear constraint says that if the toe $i$ is on the ground at both iteration $n$ and $n+1$, it should not move (slip). 

When the toe is lifted into the air, we enforce the lift height constraint and the zero force constraint:
\begin{equation}
\begin{aligned}
    p_{b,i}(z) &= H \\
    |\textbf{f}_{i}| & \leq \textbf{f}_{max}c_{i}
\end{aligned}
\end{equation}

The lift height $H$ is pre-defined. It can be a variable to deal with situations such as climbing on stairs with variable height. This will be explored in later papers.

When $c_{i}=1$, the toe is making contact with the ground. For pre-solved trajectories, we generate randomized terrain shapes discretized into multiple convex polygon regions $s=1, ..., S$, and ensure that the robot makes contact with one of the polygon region. This constraint is again mixed-integer convex. We define binary variables $z_{i,s}$ for toe $i$ and convex region $s$ such that if $z_{i,s}=1$, toe $i$ makes contact with the convex region $s$. The constraint is:

\begin{equation}
    \textbf{p}_{w,i} \in \text{Region}_{s} \ \text{if} \ z_{i,s}=1
\label{Eqn:patch}
\end{equation}

Since the toe can be lifted or make contact with one of the convex region, we enforce:
\begin{equation}
\sum_{s} z_{i,s} + (1-c_{i}) = 1
\end{equation}


\subsection{Envelope Approximation}
\label{Sec:envelope}

A standard approach to formulate the linear relaxations for bilinear constraints is through McCormick envelopes \cite{mccormick1976computability}. For a bilinear constraint $z=xy$, where $x \in [x^{L}, x^{U}]$ and $y \in [y^{L}, y^{U}]$, the McCormick envelope relaxation can be defined as:

\begin{equation}
\begin{aligned}
    z & \geq x^{L} y + x y^{L} - x^{L} y^{L}, \ z \geq x^{U} y + x y^{U} - x^{U} y^{U} \\
    z & \leq x^{U} y + x y^{L} - x^{U} y^{L}, \ z \leq x^{L} y + x y^{U} - x^{L} y^{U}
\end{aligned}
\label{Eqn:Mc}
\end{equation}

Which is the best set of linear relaxation for variable range $[x^{L}, x^{U}]$ and $[y^{L}, y^{U}]$. Multi-linear terms such as $a_{1}a_{2}...a_{n}$ usually exist in the equation of dynamics. For trilinear terms, although relaxations of higher accuracy exist \cite{meyer2004trilinear}, we simply use two McCormick envelopes. Specifically, for terms such as $a=a_{1}a_{2}a_{3}$, we first define $a_{12}=a_{1}a_{2}$ and implement Eqn. ~\eqref{Eqn:Mc} between $[a_{1}, a_{2}, a_{12}]$, then implement Eqn. ~\eqref{Eqn:Mc} again between $[a_{12}, a_{3}, a]$. We found reasonable approximation accuracy out of this approach.

Usually, this approximation accuracy is insufficient if we use the nominal variable range to generate a single McCormick envelope. The standard approach is to separate the variable range into smaller regions and implement multiple envelopes with integer variables pointing to a specific region \cite{dai2019global}. For our set of single rigid body dynamics, we identify non-convex multi-linear terms to be the $\textbf{w} \times (\textbf{I}\textbf{w})$ which is bilinear, $\textbf{r}_{b,i} \times \textbf{f}_{i}$ which is bilinear, terms in $\textbf{R}_{wb}$ which are either bilinear or trilinear, and terms from Eqn. ~\eqref{Eqn:omega} which are either bilinear or trilinear. We separate the variables into the regions as shown in Table \ref{Tab:nonconvex_regions} for envelope relaxation:

\begin{table}[h]
\caption{Segmentations of the nonconvex variables}
\begin{tabular}{@{}l|c|c@{}}
\toprule
\multicolumn{1}{c|}{variable}                               & range             & \# of regions \\ \midrule
angles {[}$\phi$, $\theta$, $\psi${]} (rad)                 & {[}-pi/2, pi/2{]} & 4             \\ \midrule
bilinear trig terms e.g. $c \theta c \psi$                  & {[}-1, 1{]}       & 4             \\ \midrule
rate of change for Euler angles e.g. $\dot{\theta}$ (rad/s) & {[}-10, 10{]}      & 16            \\ \midrule
angular velocity $\textbf{w}$ (rad/s)                       & {[}-10, 10{]}     & 16            \\ \midrule
toe position $\textbf{p}$ (cm)                              & {[}-8, 8{]}       & 4             \\ \midrule
contact force $\textbf{f}$ (N)                              & {[}-15, 15{]}     & 16            \\ \bottomrule
\end{tabular}
\label{Tab:nonconvex_regions}
\end{table}

For trigonometry terms such as $s \theta$ and $c \theta$, we implement standard piecewise linear relaxations similar to \cite{kuindersma2016optimization}.

\section{Control Implementation}\label{sec:control}

One key feature of the proposed controller is that it outputs footstep position and contact force simultaneously, as opposed to the standard force MPC approaches which typically decouple the footstep planning from the force control \cite{di2018dynamic}. One advantage of this is that the footstep planner has complete information about the force and hence can make contact decisions based on the dynamic property of the single rigid body model. We implement this controller on a quadruped robot designed and built by ourselves. The quadruped robot, named Spine Enhanced Climbing Autonomous Legged Exploration Robot (SCALER) \cite{tanaka2022scaler}, is a 18 DoF (12 joints + 6 rigid body DoF) position controlled robot for walking and wall-climbing. Each joint is actuated by a pair of dynamixel XM-430 servo motors which are designed for position control. To do force control, force-torque sensors are equipped on each end-effector. If the robot is given position and force commands simultaneously, it can track one of them perfectly, or stays somewhere in between. This controller is implemented by admittance control. Admittance controller measures external force and uses position control to track position and force profiles. Admittance controller can be formulated in task space as:

\begin{equation}
    M_{d}\ddot{\textbf{x}} + D_{d}\dot{\textbf{x}} + K_{d}(\textbf{x}-\textbf{x}_{0}) = K_{f}(\textbf{f}_{meas} - \textbf{F}_{ref})   
\end{equation}

Where $M_{d}$, $D_{d}$ and $K_{d}$ are desired mass, damping and spring coefficients. $\textbf{F}_{ref}$ is the wrench profile to be tracked. This equation can be re-written for controller design:

\begin{equation}
    \ddot{\textbf{x}} = M_{d}^{-1}( -D_{d}\dot{\textbf{x}} -K_{d}(\textbf{x}-\textbf{x}_{0}) +K_{f}(\textbf{f}_{meas} - \textbf{F}_{ref}))  
\label{Eqn:admittance_1}
\end{equation}

For admittance control, the control input is position $u$. We can use:

\begin{equation}
    u = \iint \ddot{\textbf{x}} \ dtdt
\label{Eqn:admittance_2}
\end{equation}

Equation ~\eqref{Eqn:admittance_1} ~\eqref{Eqn:admittance_2} turns the force controller into a position controller that can be directly implemented with our dynamixel servo motors.

Admittance control makes a trade-off between tracking force profile and position profile. If we set $\textbf{D}_{d}=\textbf{K}_{d}=\textbf{0}$, the position tracking is turned off and the controller becomes a force tracker. If we set $\textbf{K}_{f}=\textbf{0}$ the controller becomes a pure position tracker ignoring all measured forces. Since our trajectory contains both position and force, we feed two profiles simultaneously to the admittance controller. It can find a balance between position and force tracking.

\section{Learning for Warm Start}\label{sec:learning}

\subsection{Data-driven methods for MINLPs}
Since the problem incorporating online gait selection and non-convex dynamics is a mixed-integer nonlinear (non-convex) problem (MINLP) which is known to be difficult to solve, we implement data-driven methods to learn warm-start out of pre-solved instances and solve a simplified problem online. This idea was tested in \cite{lin2022benchmark}. We first list some of the results. There are in general two approaches to convert the MINLP into either mixed-integer programs with envelopes, or nonlinear programs with complementary constraints. The first approach, as described in section \ref{Sec:envelope}, use local convex approximations to convert the non-convex constraints into mixed-integer convex constraints. The issue with this method is that the resultant mixed-integer convex constraints are usually very slow to resolve as a precise approximation usually gives a large number of integer variables. On the other hand, one can convert the discrete variables into continuous ones with complementary constraints. Specifically, the binary variable $z \in \{0, 1\}$ may be converted into $z \in [0, 1]$ with additional constraint $z(1-z)=0$. However, this conversion tends to give a numerically difficult nonlinear optimization problem. Simply giving such a problem to NLP solvers without a good initial guess results in an extremely low feasibility rate. In both cases, pre-solving some of the problems and using learning to provide warm-starts online dramatically helps with the feasibility rate and solving speed. In the extreme case, one can take the MIP approach and learn full integer variables offline, such that the problem becomes convex online given warm-starts. Commonly, the learner will sample several integer strategies and try them one by one until a feasible solution is retrieved. Ideally, we would like to use a minimal amount of trials to get a highly optimized solution resulting in the fastest solving speed (similar to convex MPCs). This will require a training dataset of high optimality and well-performed learning. Several learning schemes have been explored \cite{cauligi2021coco, zhu2019fast}.

Other than data-driven methods, several non-data-driven methods also exist to get local optimal solutions at a decent speed such as ADMM \cite{shirai2022simultaneous}. However, ADMM can oftentimes provide highly sub-optimal solutions. With datasets of good quality, data-driven methods could provide solutions that are close to global optimal ones with even faster solving speeds. In some cases of robotics problems, global optimal are sometimes desired as they have consistent behavior across similar scenarios and are economically beneficial (for example, they complete the tasks faster and cost less energy).

\subsection{Training for locomotion with single rigid body model}

To train a learner offline, we sample the problem instances. Define the problem with a parameter $P$. Consider a quadruped robot walking subject to disturbance forces. In this case, $P$ describes the various initial conditions such as initial accelerations, velocities, angles, and angular rates caused by disturbances. In this paper, we simplify the problem by only considering flat terrain. Non-flat terrains can be considered by randomizing the terrain shape. The problem is then formulated as NLP with complementary constraints for gait variables and solved using solver IPOPT. As mentioned in the previous section, it is difficult to find feasible solutions without a good warm-start. For our problem, we use trot gait to warm-stat the binary gait variables which will be further optimized by the MIP formulation. After collecting a dataset from the NLP formulation, we use them to warm-start the equivalent MIP formulation by converting the non-convex continuous variables into integer variables. Without a warm-start, the MIP solver struggles to find any feasible solution. The MIP formulation can further improve the optimality of the dataset. One of the tricks is that we can \textit{partially} warm-start the binary variables and let the MIP solver deal with the rest of the binary variables. As the initial conditions of the problem to be solved by MIP do not necessarily match with the dataset coming from NLP solutions, using the exact binary solutions from NLP dataset may lead to infeasibility. Partially warm-start most of the binary variables can give more feasible solutions, and the MIP solving time is still fast. For the walking problem, if we warm-start the binary variables for non-convex variables and let the MIP solver deal with gait variables, the problem can be solved in a few seconds with a much more dynamic gait retunred.

One feature of MIP is that the theoretical lower bound of the optimal cost can be retrieved by relaxing the binary variables into continuous variables. The optimal primal-dual gap can be defined by the ratio of primal objective bound $z_{P}$ minus the dual objective bound $z_{D}$, or the current best objective value minus the lower bound of objective value. The gap ratio $|z_{P}-z_{D}|/|z_{P}|$ tells how optimal the current solution is. The lower this value is, the closer the current solution is to the global optimal solution. Most of the solutions from the NLP solver have an optimal gap more than $30 \%$. Since the single rigid body model is at a reasonable scale, the MIP solver Gurobi can improve the gap to less than $15\%$.

We then train a learner to provide full list of integer variables online. The general balance controller should not need the information about the terrain, and should regulate the robot body despite the change of environment. Therefore, the input of the learner is the body center of mass positions, velocities and accelerations, and the output is foot positions, body orientations and contact forces. On the other hand, the controller can be aware of the terrain if vision information is available. In this case, constraint ~\eqref{Eqn:patch} can be replaced by terrain observed by the robot. In this case, vision information can also be used as input to the learner which provides the warm-start allowing the MPC to make decisions of the foot steps on the terrain through convex optimization.


\section{Experimental Results}\label{sec:results}

We conduct several offline computation experiments using the method described in the paper. For getting an offline solution to a certain problem, we first sample the problem parameters around the targeted problem, then solve the set of problems using the nonlinear programming method with a trot gait as initial guesses. Although the chance of getting infeasible solution is high due to complementary constraints, NLPs usually finish within a few seconds. We can keep sampling until we find feasible cases. After gathering some feasible solutions using NLP, we use them to warm-start the target problem with the mixed-integer formulation. This greatly improves the solving speed of the target MIP which can oftentimes be solved to a MIP gap below $15\%$ within a reasonable time. This approach is used for solving forward walking, disturbance rejection, and large angle turning trajectories for a quadruped robot.

\begin{figure*}
		\centering
		\includegraphics[scale=0.32]{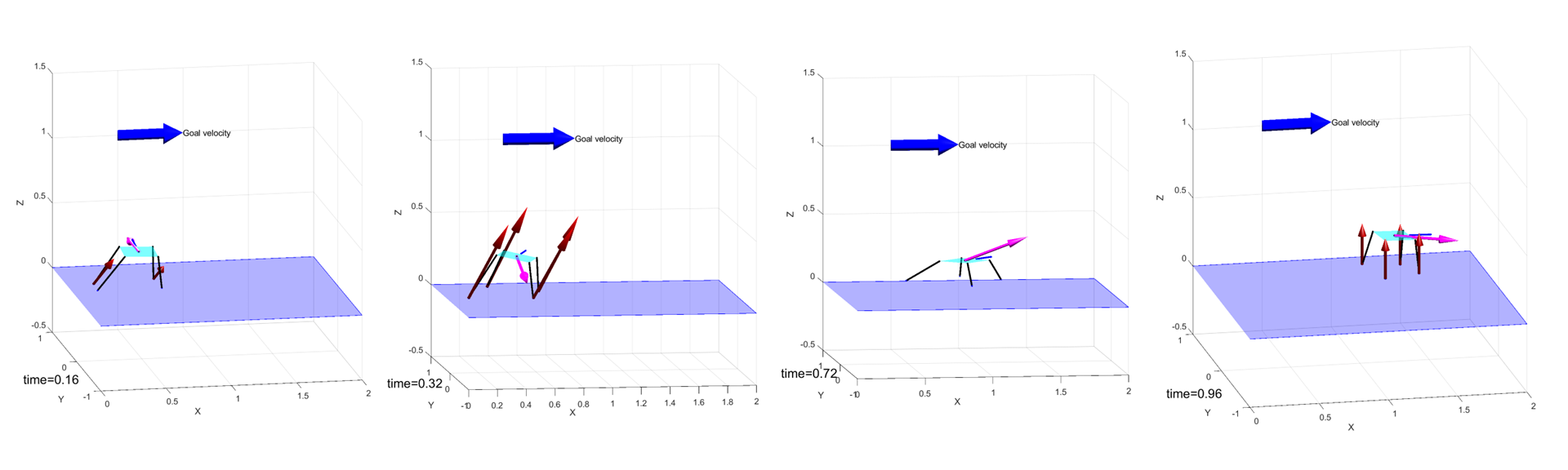}
		\caption {Large angle turning trajectory and contact forces. From left to right are snapshots at time 0.16s, 0.32s, 0.72s, and 0.96s. The red arrow represents the contact forces. The magenta arrow represents the current velocity. The short blue line pointing forwards represents the head position of the robot. Note at t=0.32, the robot tilts its body using gravity to cancel the moment from the contact forces.}
		\label{Fig:large_angle_turn_figure}
\end{figure*}

\subsection{Forward walking}
\label{Sec:forward_walking}

We first generate the control input to make the robot move forward. The objective function we use is:


\begin{equation}
\begin{aligned}
    f_{obj} &= ||\textbf{v}[i] - \textbf{v}_{ref}||^{2}_{\textbf{w}_{v}} +    ||\boldsymbol{\Theta}[i] - \boldsymbol{\Theta}_{ref}||^{2}_{\textbf{w}_{\theta}} \\
            &+ ||\textbf{p}_{z}[i] - \textbf{p}_{z,ref}||^{2}_{w_{h}} + \sum_{i} ||\textbf{p}[i+1] - \textbf{p}[i]||^{2} \\
            &+ \sum_{i, \ s \neq t} ||\textbf{f}_{s}[i] - \textbf{f}_{t}[i]||^{2}
\end{aligned}
\end{equation}

\begin{figure}[!t]
		\centering
		\includegraphics[scale=0.42]{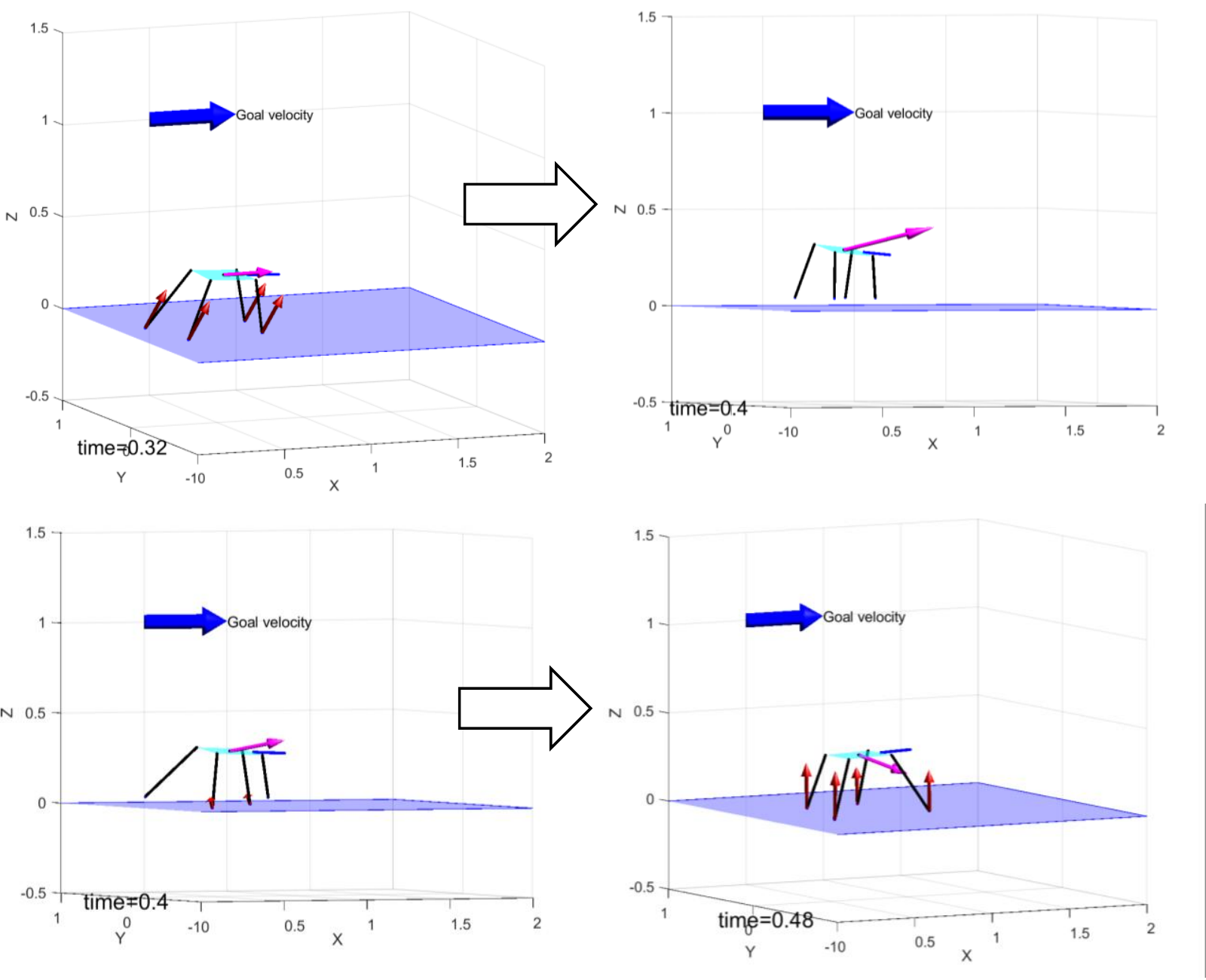}
		\caption {Forward walking trajectories and forces. The upper two figures represent the dynamic jumping trajectory from the MIP formulation which uses 4 legs to push simultaneously to maximize the forward velocity. The bottom two figures represent the trot walking trajectory from the NLP formulation which has a slower forward speed. The red arrow represents the contact forces. The magenta arrow represents the current velocity. The short blue line pointing forwards represents the head position of the robot.}
		\label{Fig:forward_walking_figure}
\end{figure}

\begin{figure}[!t]
		\centering
		\includegraphics[scale=0.12]{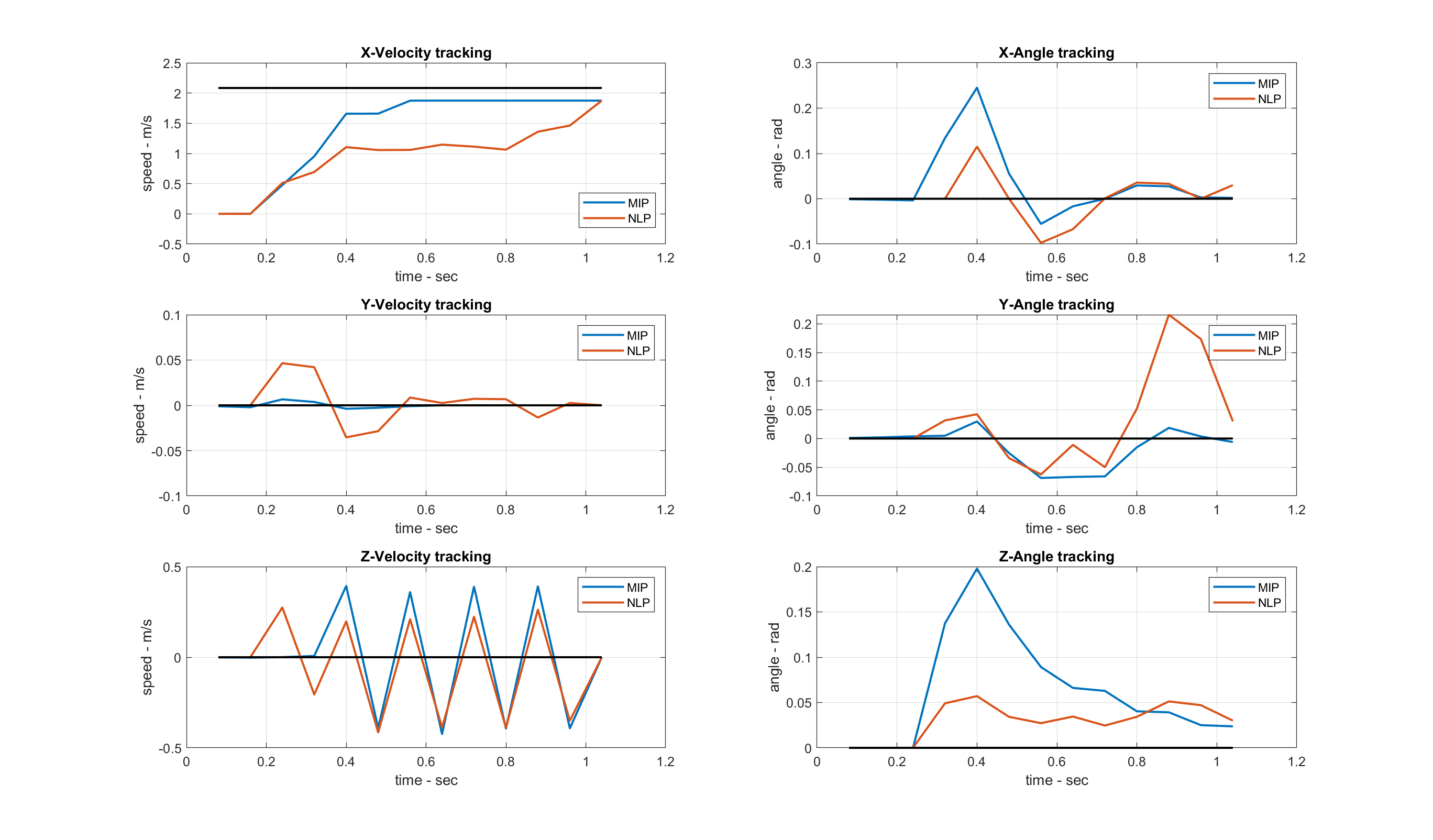}
		\caption {Velocity and angle trajectories of the forward walking task. Black lines are the tracking objectives. Orange curves come from the NLP solution. Blue curves come from the MIP solution with optimized gait. X is the forward direction and Y is the side direction. Simply changing the gait can make forward speed tracking much faster. Note that the angle tracking weights are relatively smaller, hence MIP solver may initially choose to sacrifice more angular tracking performance for better speed tracking.}
		\label{Fig:forward_walking_tracking}
\end{figure}

Where $\textbf{w}_{v}$, $\textbf{w}_{\theta}$, $w_{h}$ are the weights for velocity tracking, rotation angle tracking and body height tracking. For walking forward, we first set $\textbf{w}_{v}=[100, 100, 10]$, $\textbf{w}_{\theta}=[10, 10, 10]$, $w_{h}=10$. This set of weight favors more on the forward walking speed. As a result, the MIP solution consistently gives forward jumping gait despite that the NLP solution being handed as warm-start uses trot gait. The MIP solution, being close to the global optimal solution, significantly improves the tracking for speed as shown by Fig. \ref{Fig:forward_walking_tracking}. This is intuitively true, as jumping gaits immediately use 4 legs to push on the ground simultaneously, hence more effective in changing speed. On the other hand, if we increase the weight of $z$ velocity and position tracking to $1000$ to minimize the z direction body vibration, the optimizer gives a gait that lifts only one leg each iteration. The forward speed tracks much slower in this case. The optimized forward trajectories are shown in Fig. \ref{Fig:forward_walking_figure}.

\subsection{Disturbance rejection}
\label{Sec:Disturbance rejection}
\begin{figure}[!t]
		\centering
		\includegraphics[scale=0.12]{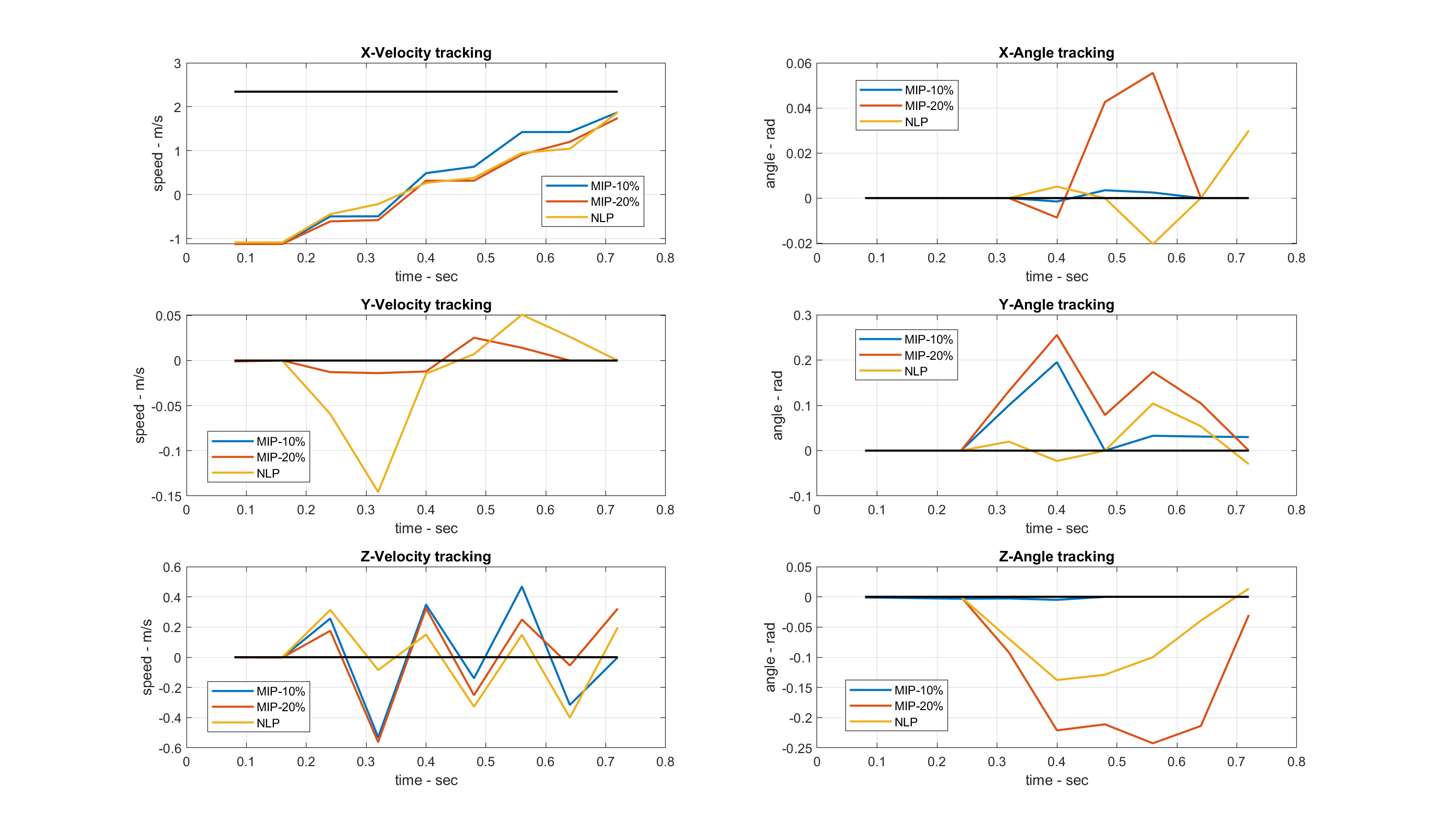}
		\caption {Velocity and angle trajectories of the disturbance rejection task. Black lines are the tracking objectives. The 20\% gap MIP solution takes 30 min to solve, and the more optimal 10\% gap MIP solution takes 45 min. X is the forward direction and Y is the side direction. As the 20\% gap MIP solution still does unnecessary motions such as out of plane (Z) rotation, the 10\% gap MIP solution almost does not do any unnecessary motions.}
		\label{Fig:Disturbance_rejection_tracking}
\end{figure}

In this experiment, we test the controller performance when the objective is moving forward as in the previous section, but there is a large initial backward speed as a disturbance. If the backward speed is relatively smaller, The controller commands all the legs to be on the ground for a few iterations, then resumes the forward jumping trajectory. In this situation, all legs create forward forces to cancel the back speed, effectively serving as a brake. If the backward speed is relatively larger, the leg kinematics will be infeasible due to the no-slip constraint as the body quickly goes backward. In this case, the controller generates a back trot gait for a few iterations, then resumes the forward jumping gait. The tracking performance is shown in Fig. \ref{Fig:Disturbance_rejection_tracking}.

\subsection{Large angle turning}
\begin{figure}[!t]
		\centering
		\includegraphics[scale=0.12]{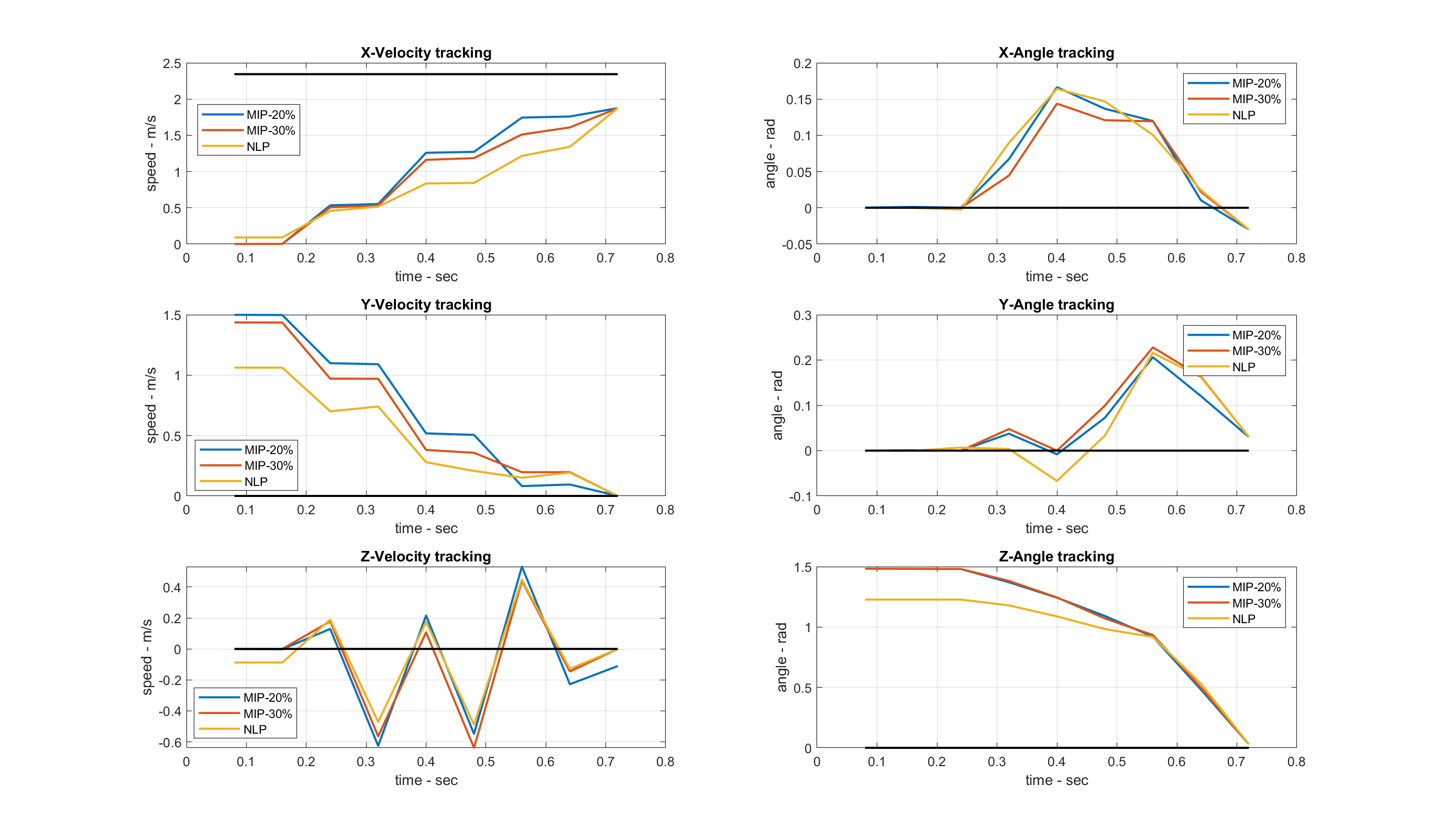}
		\caption {Velocity and angle trajectories of the large angle turning task. Black lines are the tracking objectives. The 30\% gap MIP solution takes 30 min to solve, and the more optimal 20\% gap MIP solution takes 90 min. X is the forward direction and Y is the side direction. Although the NLP solution has much better initial conditions (we cannot find a feasible solution with the exact initial condition to the MIPs), both the MIP solutions quickly catch up and track the objectives faster.}
		\label{Fig:large_angle_turn_tracking}
\end{figure}

In this section, we validate the controller performance to make large angle turns. This serves to verify the controller's ability to select gait and make large orientation changes simultaneously. The initial condition given is $1.8 m/s$ forward while the target tracking speed is $1.8 m/s$ but at a $90 deg$ angle to the side. The NLP solution with initial guesses of trot gait remains using the trot gait to perform the turning. However, the trajectory improved by the MIP will generate a jumping-type of gait with a jump first to cancel the forward velocity. It then simultaneously walks sideways and rotates the body using a non-traditional gait. The velocity and angle tracking plots are shown in Fig. \ref{Fig:large_angle_turn_tracking}, and the snapshots of the trajectory are shown in Fig \ref{Fig:large_angle_turn_figure}. Intuitively, the more optimized MIP solution makes the motion much more dynamic. Note that when the robot breaks to cancel the forward velocity (iteration 4, time=0.32s), it needs to generate a large breaking force on its toe. The robot shifts the center of mass backward using the moment from gravity to cancel the moment from the braking force, so a larger braking force can be used. 

\subsection{Approximation accuracy}

In this section, we show the approximation accuracy using McCormick envelopes for bilinear and trilinear terms. Since we use envelopes to locally convexify the constraints, it is important to ensure reasonable approximation accuracy. There are several types of non-convex terms: bilinear trig multiplication terms such as $sin(\theta_{0})cos(\theta_{1})$, bilinear angular velocity multiplication terms such as $w_{0}w_{1}$, bilinear moment terms such as $p_{x}f_{y}$, trilinear trig multiplication terms such as $sin(\theta_{0})sin(\theta_{1})cos(\theta_{2})$, trilinear trig and Euler angular rate multiplication terms such as $sin(\theta_{2})cos(\theta_{1})\dot{\theta}_{0}$. We average the approximation accuracy of those terms across the trajectory for a 200 point dataset using the following equation:

\begin{equation}
    error_{x} = \frac{|x_{approx}-x_{true}|}{max(|x_{approx}|, |x_{true}|)}
\end{equation}

\begin{figure}
		\centering
		\includegraphics[scale=0.28]{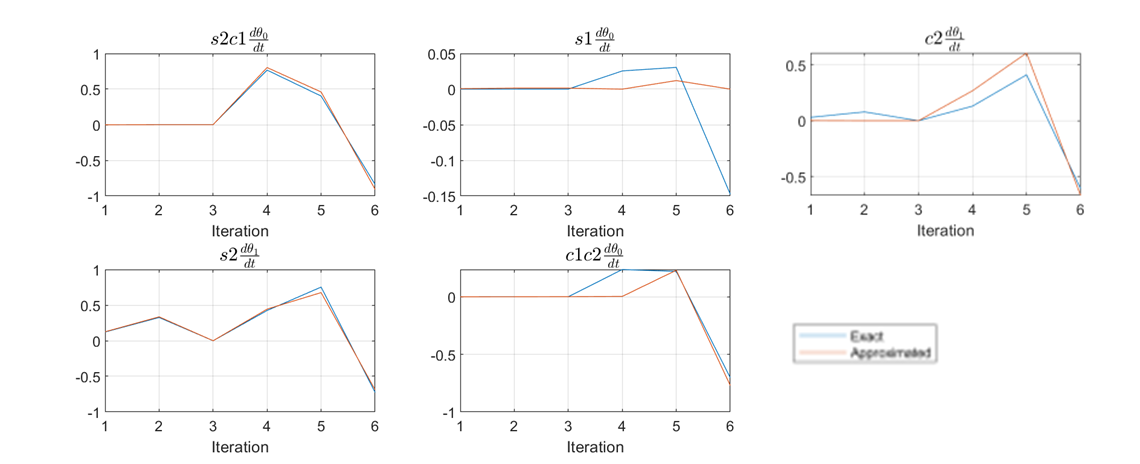}
		\caption {Approximation accuracy along a typical trajectory.}
		\label{Fig:trilinear_error}
\end{figure}

The results are listed in Table \ref{Tab:approximation_accuracy}. An actual trajectory of approximated and true values for a few non-convex terms is shown by Fig. \ref{Fig:trilinear_error}. The results show that the approximated values clearly resemble the trend of change in the actual values, and the accuracy is generally around $10\%$ or less. We note that since the model may not be perfect, a rough approximation leaves some room for tuning on actual hardware which is sometimes favored.

\begin{table}[H]
\centering
\caption{Average approximation accuracy}
\begin{tabular}{@{}c|c|c|c|c|c@{}}
\toprule
term  & $s\theta_{0}c\theta_{1}$ & $s\theta_{0}s\theta_{1}c\theta_{2}$ & $s\theta_{2}c\theta_{1}(\dot{\theta}_{0})$ & $w_{0}w_{1}$ & $p_{x}f_{y}$ \\ \midrule
error & 10.84\%                  & 12.38\%                             & 2.86\%                                     & 11.73\%      & 7.81\%       \\ \bottomrule
\end{tabular}
\label{Tab:approximation_accuracy}
\end{table}

\subsection{Dataset collection and hardware implementation}

\begin{figure*}
		\centering
		\includegraphics[scale=0.24]{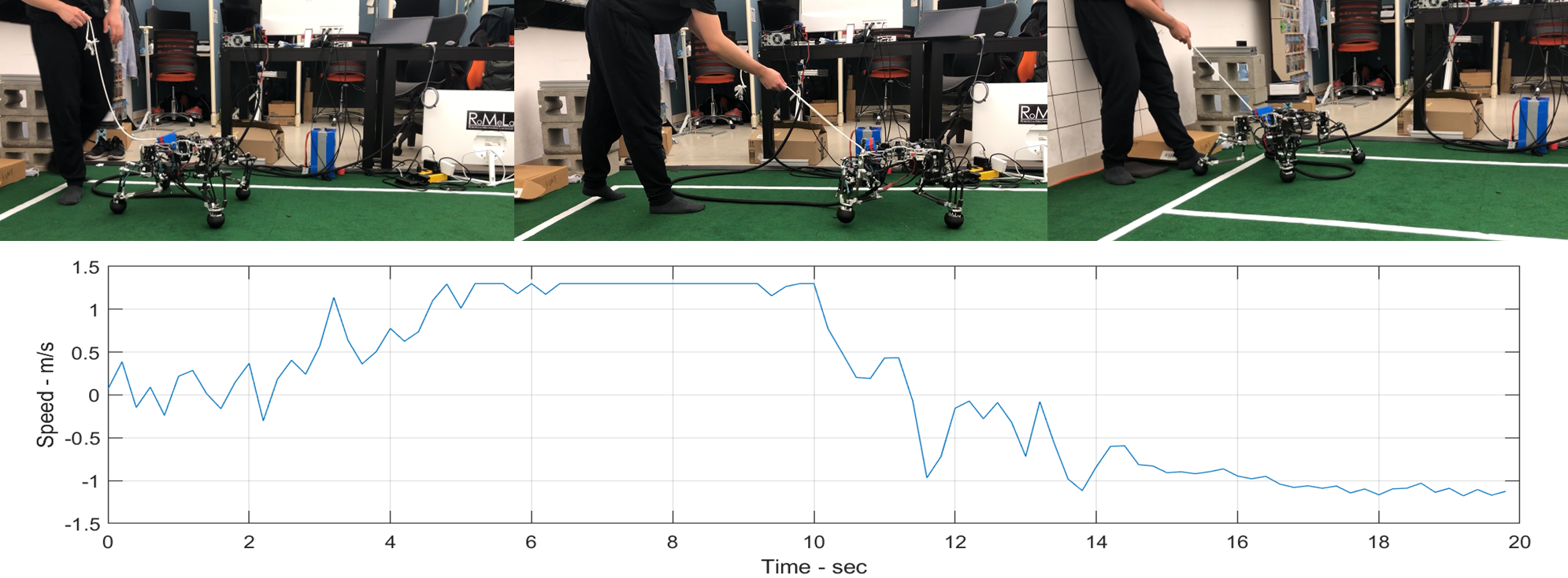}
		\caption {Hardware experiment for walking. Up: 3 snapshots of the robot walking forward using the proposed framework and changing gait when dragged back. Down: estimated forward speed from state estimation.}
		\label{Fig:hardware_experiment}
\end{figure*}

We implemented our proposed control framework on a position-controlled quadruped robot SCALER. The experiment is a forward walking task similar to what is described in section \ref{Sec:forward_walking} and \ref{Sec:Disturbance rejection}, designed to observe online gait change due to MPC results given the state of the robot. Since this robot is position controlled, we use the admittance control law with force torque sensors on its feet to track the planned position and force profile. This same controller is used throughout the trajectory. We collected 110 trajectories with 720 ms long, 9 iterations each, that begin with randomized initial conditions of $vx \in [-1.5, 1.5]$ and $ax \in [-15, 15]$. Those trajectories give information on how the robot can reach the goal velocity from different initial conditions in the most optimal approach. Since our robot cannot track the jumping trajectory, we force the optimizer to have at least 2 contact points on the ground for each iteration. The MPC plan 5 iterations ahead (400 ms) for the robot to follow the trajectories and reach the goal. Along the full trajectory, we segment out 5 iteration sections, 220 in total, and use them as data for the learner. Our previous tests on smaller scale problems \cite{lin2021reduce, hauser2016learning} show that the K-nearest neighbor method works decently well on problems of this scale, hence is used in this experiment. Typical force MPC algorithms only need to know the current state to plan ahead. This is because there is a pre-generated gait pattern from a high-level planner. However, for our hybrid MPC, the gait and forces are both generated online. To reason the gait which is of slower changing frequency, the learner needs to have information of a short history of the trajectory to decide the next lifting leg. Otherwise, the learner may keep on lifting the same set of legs. We feed the current forward velocity, forward acceleration, and the toe positions in the previous 2 iterations to the KNN learner which provides the integer variables in the next 5 iterations. With the integer variables provided, convex optimization is solved for the actual contact positions, forces, body orientations, and corresponding speed and acceleration quantities. We use OSQP solver \cite{stellato2020osqp} for online solving. One feature of KNN is that it can provide several candidates of integer variables. If one fails, the solver can try the next one until one succeeds. For this problem, as long as the input to KNN is within the region of the randomization regime, it usually takes only 1 trial to find a feasible solution. The average solving speed is 19 ms on an Intel Core i7-12800H laptop or 53 Hz. State estimation \cite{bloesch2013state} that fuses encoder and IMU information is used to estimate the forward velocity used by KNN.

The hardware experiment begins with the robot making a step forward that gains some speed, which is fed back to KNN such that the robot proceeds to follow the trajectory. The human operator then manually pulls the robot backward. With the sensed negative speed, the robot then stops trotting forward but keeps all its toes on the ground to maximize the braking force as described in section \ref{Sec:Disturbance rejection}. The process is shown in the upper part of Fig. \ref{Fig:hardware_experiment}. The estimated forward velocity is shown in the lower part of the same figure. 

\subsection{Solving speed}
\label{Sec:solving_speed}
For mixed-integer nonconvex problems such as locomotion planning with a single rigid body, directly solving the MIP formulation with envelope constraints is mostly infeasible. The solver can sometimes take days before it can find even one feasible solution. First sampling the problem parameters around the targeted problem and solving a small set of problems in NLP formulation, then using them to partially warm-start the target problem in MIP formulation dramatically speeds up the MIP solving speed. For our single rigid body problem, the MIP solver with warm-start tends to make quick progress to minimize the primal-dual gap in the next 30 minutes to 1 hour, until it reaches around 15\% and slows down again. For the disturbance rejection task, the solver takes 30 minutes to decrease the bound to 20\%, and 45 minutes to further decrease it to 10\%. For the large angle turning task, it takes 30 minutes to decrease the bound to 30\%, and 90 min to further decrease it to 20\%. In many cases, this process already makes the solution significantly more optimal than the original solutions from NLP solvers. If the goal is to optimize the gait, the MIP solver takes seconds to finish while keeping the large angle rotation trajectories from NLP solvers. After MIP solving, the more optimal solutions can be added to the dataset which makes the following process even faster.

Table \ref{Tab:problem_size} gives the number of variables, constraints, and solving time for both the offline and online optimizations. The online MPC with KNN learned integer variables can run more than 50 Hz on the OSQP solver which is sufficient for hybrid MPC. In comparison, solving the same problem with NLP formulation on the IPOPT solver is significantly slower. The commercialized KNITRO solver may run faster than IPOPT but we expect it not to be able to catch the convex MPC solving speed.

\begin{table*}[]
\centering
\caption{Problem sizes and solving speeds}
\begin{tabular}{lcccc}
\hline
\multicolumn{1}{c}{}       & Offline TO (NLP)                                                                               & Offline TO (MIP)                                                        & Online MPC    & Online NLP benchmark                                                                                 \\ \hline
\# of iterations           & 9 (720 ms)                                                                                     & 9 (720 ms)                                                              & 5 (400 ms)    & 5 (400 ms)                                                                                           \\ \hline
\# of continuous variables & 1143                                                                                           & 80321                                                                   & 34549         & 579                                                                                                  \\ \hline
\# of binary variables     & N.A.                                                                                              & 976                                                                     & 488           & N.A.                                                                                                    \\ \hline
\# of constraints          & 2327                                                                                           & 103430                                                                  & 44478         & 1115                                                                                                 \\ \hline
Avg. solving time          & \begin{tabular}[c]{@{}c@{}}1.34 s (when feasible with \\ trot gait initial guess)\end{tabular} & See section \ref{Sec:solving_speed} & 19 ms (53 Hz) & \begin{tabular}[c]{@{}c@{}}498 ms (2 Hz, when feasible with \\ trot gait initial guess)\end{tabular} \\ \hline
Solver                     & IPOPT                                                                                          & Gurobi                                                                  & OSQP          & IPOPT                                                                                                \\ \hline
\end{tabular}
\label{Tab:problem_size}
\end{table*}










\section{Discussion and Future Works}\label{sec:discussion}

This paper proposes a method to learn offline solutions with a single rigid body model using mixed integer non-convex optimization and uses those solutions to run hybrid model predictive control online. The results show that with warm-starts, the MIP formulation can be solved close to the global optimal solution. With offline learned integer variables, the online problem can be solved using a convex optimization solver at a real-time control speed. We implemented the hybrid MPC for 2D walking with gait change. Due to the limitation of this hardware, we cannot implement any jumping motion or motions incorporating high bandwidth force control. Thus, the hardware implementation for motions such as large-angle turning is left as the next work.

Since MIP formulations can generate nearly global-optimal behaviors. It is interesting to observe how the MIP solver makes decisions under various scenarios and understands how they make sense. As we can generate datasets for optimal behaviors, we are also seeking to reuse them for tasks under different situations, different robots, or problems of different sizes to maximize the usage of data.

\section{Acknowledgements}
We would like to thank Junjie Shen, Gabriel Fernandez, Yuki Shirai for helpful discussions.












\bibliographystyle{IEEEtran}
\bibliography{references}

\end{document}